\documentclass{article}

\usepackage{PRIMEarxiv}

\usepackage[utf8]{inputenc} 
\usepackage[T1]{fontenc}    
\usepackage{hyperref}       
\usepackage{url}            
\usepackage{booktabs}       
\usepackage{amsfonts}       
\usepackage{nicefrac}       
\usepackage{microtype}      
\usepackage{lipsum}
\usepackage{fancyhdr}       
\usepackage{graphicx}       
\graphicspath{{media/}}     

\usepackage{siunitx}
\usepackage{caption}
\usepackage{subcaption}
\usepackage{multirow}
\usepackage{hyperref}
\usepackage{tikz}
\usetikzlibrary{positioning, shapes.geometric}
\usepackage{amssymb}
\usepackage{bbm}
\usepackage{booktabs}
\usepackage{tabularx}
\usepackage{amsmath}

\title{Autonomous Robotic Navigation for Endovascular Brain–Computer Interface Access
}

\author{\normalfont
Harry Robertshaw\textsuperscript{1,2},
Weijie Qi\textsuperscript{2},
Nikola Fischer\textsuperscript{1},
Alejandro Granados\textsuperscript{1},
Thomas C. Booth\textsuperscript{1,3},
Sam E. John\textsuperscript{2},\thanks{Corresponding author: sam.john@unimelb.edu.au}
\\[1em]
\textsuperscript{1}School of Biomedical Engineering \& Imaging Sciences, Kings College London, London, UK\\
\textsuperscript{2}Department of Biomedical Engineering, The University of Melbourne, Parkville, Victoria, Australia\\
\textsuperscript{3}Department of Neuroradiology, King's College Hospital, London, UK\\
}

\begin{document}
\maketitle

\begin{abstract}
    Endovascular brain--computer interfaces (BCIs) avoid craniotomy but require precise device delivery through anatomically variable cerebral veins. This work presents the first demonstration of \textit{in vitro} autonomous robotic navigation for endovascular BCI access in the cerebral venous system. Soft Actor-Critic controllers were trained \textit{in silico} for two sequential tasks spanning the right internal jugular vein to the superior sagittal sinus, using geometric augmentation of one training anatomy. Navigation was evaluated in a training anatomy and an anatomically unseen hold-out model over 250 \textit{in silico} episodes and five fluoroscopy-guided \textit{in vitro} robotic runs per task–anatomy condition, comprising 1,000 simulated episodes and 20 physical runs overall. Task recurrent predictors were also evaluated for online identification of impending navigation failure. \textit{In silico} success rates for Tasks A and B were 85.6\% and 98.4\% in the training anatomy and 42.0\% and 91.6\% in the hold-out anatomy, respectively. Fourteen of 20 physical runs were successful (70\% overall), including 80\% success for Task B in the hold-out phantom. \textit{In silico} the predictors detected 99.3--100.0\% of failures with false-alarm rates of 0.8--6.7\%. During \textit{in vitro} evaluation, predicted risk increased before failed episodes, but elevated probabilities during some successful runs showed reduced calibration after transfer. These results demonstrate the feasibility of autonomous cerebral venous access and show how online failure prediction could support human oversight, while also identifying anatomical generalization and sim-to-real calibration as priorities before preclinical translation.
\end{abstract}


\section{Introduction}

    Brain–computer interfaces (BCIs) offer a route to restoring communication and control through digital devices for individuals with severe motor impairment arising from conditions such as amyotrophic lateral sclerosis (ALS), brainstem stroke, or high cervical spinal cord injury, for whom conventional assistive technologies are frequently insufficient~\cite{Chaudhary2016, NicolasAlonso2012}. Cortical BCIs achieve this by decoding motor intent directly from neural activity, bypassing the damaged efferent pathways that would otherwise carry a movement command from brain to muscle, and translating that decoded intent into control of an external device~\cite{Jackson2012}.
    
    The majority of BCIs validated in human trials to date have relied on electrode arrays implanted via craniotomy~\cite{PatrickKrueger2025}. This could be by penetrating the cortical surface as intracortical arrays, or resting upon it as subdural or epidural electrocorticography (ECoG) grids. These approaches offer excellent spatial and temporal signal fidelity, but craniotomy carries surgical risk, and the chronic foreign-body response provoked by direct cortical contact, including glial scarring and encapsulation, can progressively degrade signal quality over the working life of the implant~\cite{Salatino2017, Campbell2018}.
    
    Endovascular BCIs offer an alternative to open cranial surgery. The Stentrode (Synchron, Brooklyn, NY, USA) is a stent-mounted electrode array delivered via the jugular vein and deployed within the superior sagittal sinus (SSS) adjacent to primary motor cortex. This approach repurposes catheter-based delivery techniques established in interventional cardiology and neurointervention for the placement of a chronic neural interface. Endovascular BCIs enable neural recording via cerebral veins, avoiding craniotomy, with signal quality similar to subdural arrays and minimal vascular complications reported to date \cite{Ognard2025}. The SWITCH and COMMAND feasibility studies have now extended to ten participants implanted for periods of up to five years, and have shown that vascular ECoG can be recorded safely and with long-term signal stability, although considerable inter-participant variability in recorded motor signal strength has been observed~\cite{Schone2025,Chetty2025}.
    
    Functional performance of an endovascular BCI depends on the relationship between electrode location and task-relevant cortex. Clinical data indicate that greater primary motor cortex coverage is associated with stronger recorded motor signals from the Stentrode~\cite{Schone2025}, meaning achieving reproducible access to patient-specific venous targets is an important prerequisite for reliable endovascular BCI implantation. This is challenging because cerebral venous anatomy varies between individuals and the relevant vessels can be tapered and tortuous~\cite{Oxley2016}. Additionally, computational modeling suggests that vessel tapering and deformation alter the local hemodynamic environment around stent-mounted neural interfaces~\cite{Qi2024}.

    Endovascular BCI placement remains a manually performed neurointervention. Under fluoroscopic guidance, the operator repeatedly translates and rotates the guidewire and catheter to reach a prespecified vascular target. Similar to other endovascular procedures, manual navigation is dependent on operator skill, while repeated fluoroscopic imaging exposes operators to occupational radiation~\cite{Robertshaw2023}. Autonomous robotic navigation could reduce operator burden and, in the case of endovascular BCIs, improve the reproducibility of access to prespecified venous targets, independent of an individual operator's skill.
    
    Previous research into the autonomous navigation of catheters and guidewires has demonstrated \textit{in silico} and \textit{in vitro} feasibility across cardiac interventions, mainly by using reinforcement learning (RL) to automate device control~\cite{Robertshaw2023}. Recently, autonomous thrombectomy navigation has been investigated using multi-agent and world-model-based RL approaches validated across unseen patient vasculatures \textit{in silico} and fluoroscopy-guided vascular phantoms \textit{in vitro}~\cite{Robertshaw2026ral,Robertshaw2026iros}.
    
    Despite these studies, current literature has focused predominantly on the aortic arch and coronary vasculature~\cite{Robertshaw2023}, with only a small number of studies extending into the cerebral vasculature, and those exclusively in the context of the arterial circulation~\cite{Robertshaw2025ipcai, Barnes2025}. 

    To our knowledge, no study has addressed autonomous navigation of the cerebral venous system. Beyond demonstrating navigation performance, recent work has increasingly emphasized human-supervised autonomy embedded within existing interventional workflows~\cite{Robertshaw2026jaha, Granados2026}. Under this framing, autonomous navigation should be accompanied by online monitoring that identifies when system behavior is becoming unreliable and when human intervention may be required. This motivates the combination of autonomous venous navigation and online failure prediction investigated in this study.

    The primary aim of this paper was to demonstrate that the autonomous navigation of the cerebral venous system is possible both \textit{in silico} and \textit{in vitro}. Specifically, we contribute:

    \begin{itemize}
        \item The first demonstration of autonomous robotic navigation for cerebral venous access, including fluoroscopy-guided robotic navigation from the right internal jugular vein toward the superior sagittal sinus.
        \item Sim-to-real evaluation of controllers trained entirely \textit{in silico}, including evaluation in an anatomically unseen cerebral venous \textit{in vitro} model.
        \item A task-specific recurrent failure-prediction framework for online estimation of impending navigation failure, evaluated \textit{in silico} and transferred to the physical testbed to investigate its potential for human-supervised autonomy.
    \end{itemize}

\section{Methods}

    \subsection{Navigation tasks}

        We decomposed the access route into two sequential navigation subtasks, following a similar strategy used for arterial endovascular navigation~\cite{Robertshaw2025miccai}. It has also been shown that these sub-tasks can be combined to complete long continuous navigation tasks \textit{in silico} and \textit{in vitro}~\cite{Robertshaw2026ral}.

\begin{figure}[tb]
            \centering
            \includegraphics[width=0.8\linewidth]{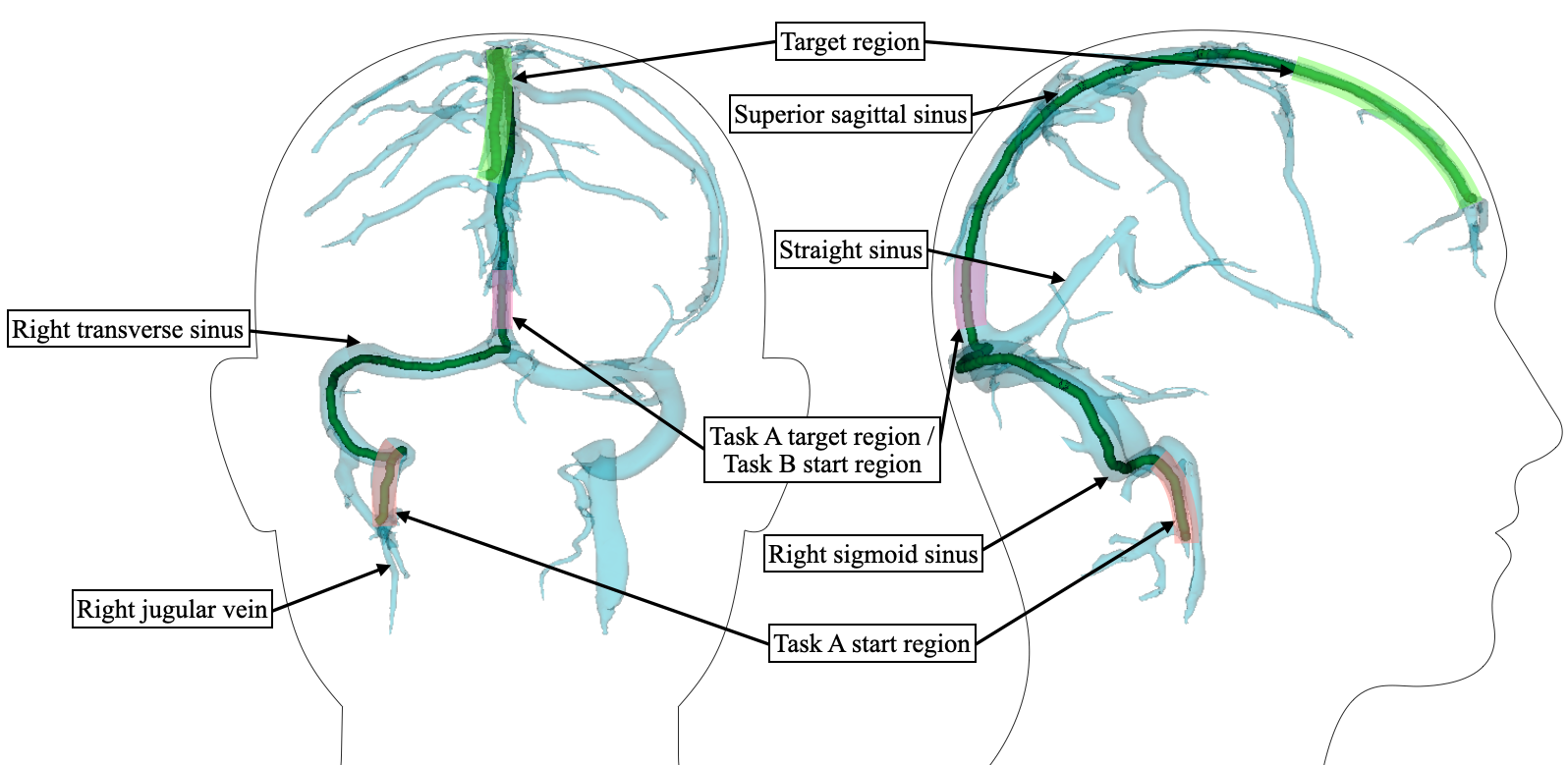}
            \caption{ Labeled anatomy for navigation tasks A and B, with task start and target region highlighted.}
            \label{fig:tasks}
        \end{figure}

        Here, navigation task A involves navigation from the superior right jugular vein to the inferior SSS, representing the destination for a stent that targets the cerebellum or visual cortices. Task B then starts at the inferior SSS and navigates to the most superior aspect of the SSS, representing the destination of a stent for motor decoding or somatosensory decoding. Both navigation tasks are shown in Fig.~\ref{fig:tasks}, and in Table~\ref{tab:bci_tortuosity} where it can be seen that Task A has a longer length and higher tortuosity than task B. Together, Tasks A and B approximate successive segments of the vascular access route used for Stentrode implantation, from the right internal jugular vein to the SSS adjacent to primary motor cortex.

        \begin{table}[tb]
            \centering
            \caption{Centreline length and tortuosity for each anatomy and task. Length was calculated as the sum of the Euclidean distances between consecutive centreline points. Tortuosity was calculated as the centreline length divided by the straight-line distance between the first and last centreline points; a value of 1 indicates a straight path.}
            \label{tab:bci_tortuosity}
            \begin{tabular}{llcc}
                \hline
                Anatomy & Task &
                Length (mm) & Tortuosity \\
                \hline
                Training & A  & 224.6 & 1.72 \\
                Training & B & 145.4 & 1.49 \\
                Hold-out & A  & 211.9 & 1.73 \\
                Hold-out & B & 161.3 & 1.39 \\
                \hline
            \end{tabular}
        \end{table}

        For all navigation tasks, a target was randomly sampled from a set of points within the task target region, while a catheter and guidewire were navigated to it from a randomly sampled point in the task start region. 

    \subsection{Dataset}\label{dataset}

        Two publicly available 3D vascular STL models derived from MRI venography were used in this study. The models, \textit{CNS Venography 3D SR Nevit Dilmen} and \textit{3DPX-003440 CNS Venography Nevit Dilmen}, were created by Nevit Dilmen and made available through Wikimedia Commons under a Creative Commons Attribution-ShareAlike 3.0 (CC BY-SA 3.0) license~\cite{DilmenCNSVenography,Dilmen3DPX}. The models contain the cerebral venous structures required for the navigation tasks, including the SSS and associated venous structures relevant to endovascular BCI implantation.

        The source STL models were subsequently modified for use in the navigation environment. They were imported into 3D Slicer~\cite{Fedorov2012}, where vessel centrelines were extracted. Local vessel radii were calculated at each centreline point and used to reconstruct a parameterized representation of the vascular lumen. This representation enabled generation of a smooth surface mesh while preserving the underlying vascular topology and vessel diameter distribution. The resulting surface meshes were subsequently imported into the Simulation Open Framework Architecture (SOFA, v23.12), where they were used for autonomous navigation training and evaluation~\cite{Faure2012}.

        As only two anatomically distinct venous models were available, one was used for training and one as a hold-out anatomy. Additional geometric augmentation was employed during RL training to increase morphological diversity and reduce overfitting to a single anatomy~\cite{Robertshaw2026iros}. Augmentations were applied directly to the extracted centreline representation prior to mesh generation. These included global scaling, local stretching and compression of vessel segments, rotation, and spatial perturbation of centreline control points while preserving vessel connectivity. Vessel radii were transformed consistently with the modified centrelines to maintain realistic lumen geometry. Centreline lengths were randomly scaled by $\pm20\%$, vessel radii by $\pm20\%$, and rotational perturbations of up to $30^{\circ}$ were applied.

        The \textit{in silico} augmented vascular models were used during training to expose the navigation policy to a wider range of vessel geometries, diameters, and curvatures than available in the original source models. This approach aimed to improve robustness to anatomical variability and facilitate generalization to previously unseen vascular anatomies. 

    \subsection{Testbed}

        \subsubsection{In silico}

            The open-source simulated EndoVascular Environment (stEVE) framework was used to build an \textit{in silico} environment~\cite{Karstensen2025}. stEVE utilizes the BeamAdapter plugin for SOFA, which was used to model the vasculatures and the devices used for navigation~\cite{Wei2012}. The devices modeled were a $0.038"$ J-tip guidewire (AngioDynamics, Inc., Latham, NY, USA), and a $0.070"$ multipurpose (MP) catheter (Penumbra, Inc., Alameda, CA, USA).

            A single-tracking (guidewire tracking only) method was employed~\cite{Robertshaw2024}. The simulation's fidelity to real-world device behavior was achieved through iterative fine-tuning of parameters such as the friction between the device and vessel wall and the device's stiffness. Such methodology has previously allowed domain adaptation of autonomous navigation agents from \textit{in silico} to \textit{in vitro} models and from \textit{in silico} to \textit{ex vivo} of porcine liver vasculatures~\cite{Karstensen2025,Karstensen2022,Robertshaw2026ral}.

        \subsubsection{In vitro}

\begin{figure}[tb]
                \centering
                \includegraphics[width=0.6\linewidth]{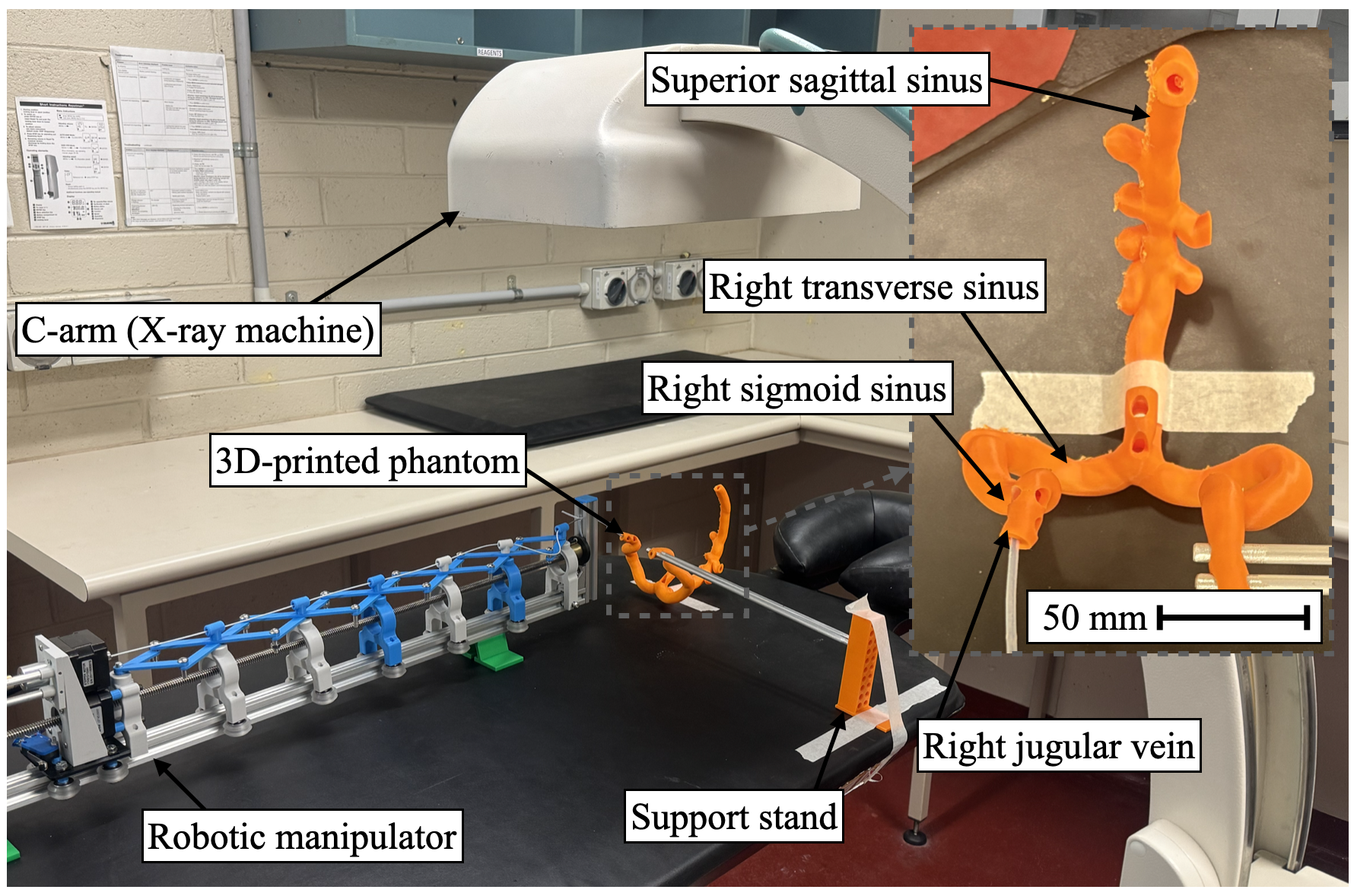}
                \caption{Labeled \textit{in vitro} testbed, with inset of 3D vascular phantom and labeled anatomy.}
                \label{fig:phantom}
            \end{figure}

            The \textit{in vitro} testbed used in this study consisted of an in-house 3D-printed phantom fabricated from polylactic acid (PLA), as shown in Figure~\ref{fig:phantom}. Both of the vascular models described in Section~\ref{dataset} were printed at original scale. A mobile C-arm (Ziehm Vision, Ziehm Imaging,  Nuremberg, Germany) was used to provide fluoroscopic images as an input to a device tip-tracking algorithm that extracted the necessary coordinates for the state-based RL algorithm~\cite{Eyberg2022}. A $0.038"$ J-tip guidewire (AngioDynamics, Inc., Latham, NY, USA), and a $0.070"$ MP catheter (Penumbra, Inc., Alameda, CA, USA) were used for experiments.

\begin{figure}[tb]
                \centering
                \includegraphics[width=0.6\linewidth]{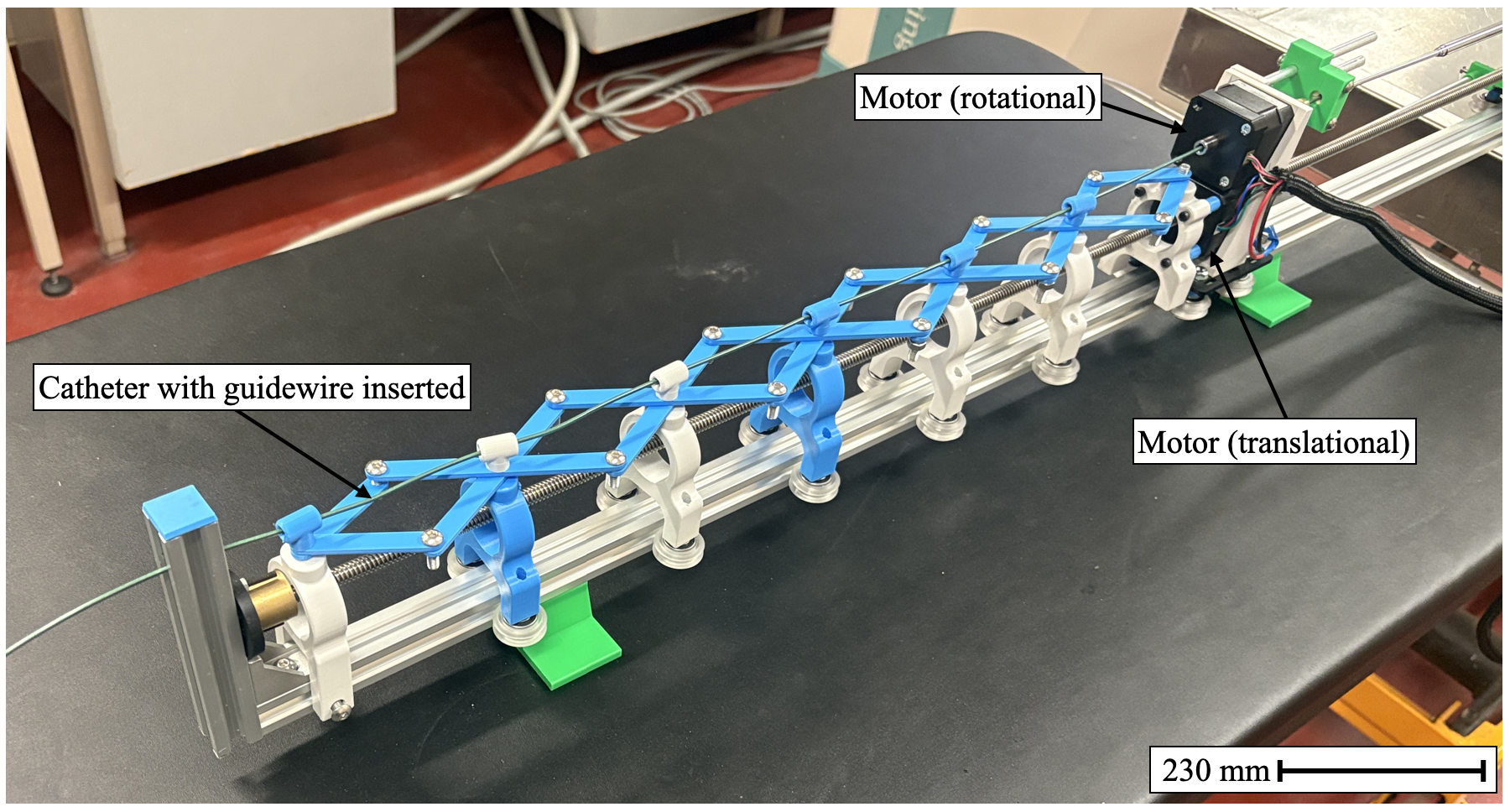}
                \caption{Robotic manipulator used for \textit{in vitro} evaluation.}
                \label{fig:robot}
            \end{figure}

            A robotic manipulator (Fig.~\ref{fig:robot}) was used to translate and rotate the catheter and guidewire independently~\cite{Robertshaw2026ral}. The system features a modular design with independent actuation carriage units, each capable of two degrees of freedom (rotation and translation), placed on a single lead screw rod, which was fixed on a frame made from aluminum struts. Direct-drive hollow shaft stepper motors were used with a leadscrew nut to translate the actuation carriage unit with an instrument along the length of the actuation system. Furthermore, the same stepper motor type was deployed to rotate the instruments via small drill chuck couplings.

    \subsection{Controller architecture}

        \subsubsection{Observation and action spaces}

            Navigation was formulated as a partially observable Markov decision process. At control step $t$, the recurrent policy selected an action from the history of observations. The current observation $\mathbf{o}_t$ comprised the current guidewire positions $\mathbf{p}_t^\top$, the previous guidewire positions $\mathbf{p}_{t-1}^\top$, episode target $(\mathbf{g}^{(e)})^\top$, and previous action $\mathbf{a}_{t-1}^\top$ as in \eqref{eq:observation}.
            
            \begin{equation}
                \begin{aligned}
                    \mathbf{o}_t &=[\mathbf{p}_t^\top,\mathbf{p}_{t-1}^\top,
                                     (\mathbf{g}^{(e)})^\top,\mathbf{a}_{t-1}^\top]^\top,\\
                    \mathbf{p}_t &=[x_{d,1,t},y_{d,1,t},\ldots,
                                     x_{d,3,t},y_{d,3,t}]^\top .
                \end{aligned}
                \label{eq:observation}
            \end{equation}
            
            Here, $\mathbf{p}_t$ contains the image-plane coordinates of three equally spaced points along the distal guidewire, with $(x_{d,1,t},y_{d,1,t})$ at the tip. The controller received neither the fluoroscopic image nor the vessel geometry. The normalized action $\mathbf{a}_t$ and its mapping to physical translational $\hat{v}_{j,t}$ and rotational $\hat{\omega}_{j,t}$ velocities were given as \eqref{eq:action}.
            
            \begin{equation}
                \begin{aligned}
                    \mathbf{a}_t={}&[\hat{v}_{\mathrm{gw},t},\hat{\omega}_{\mathrm{gw},t},
                    \hat{v}_{\mathrm{cat},t}, \hat{\omega}_{\mathrm{cat},t}]^\top
                    \in[-1,1]^4,\\                    
                    v_{j,t}=&40\hat{v}_{j,t}\;\mathrm{mm\,s}^{-1},
                    \qquad
                    \omega_{j,t}={}180\hat{\omega}_{j,t}\;{}^{\circ}\mathrm{s}^{-1},\\[-2pt]
                    &j\in\{\mathrm{gw},\mathrm{cat}\}.
                \end{aligned}
                \label{eq:action}
            \end{equation}

        \subsubsection{Soft Actor--Critic}
            
            Controllers were trained with Soft Actor--Critic (SAC), an off-policy maximum-entropy actor--critic algorithm~\cite{Haarnoja2018}, using the recurrent architecture described in~\cite{Karstensen2025}. An LSTM observation embedder was followed by separate policy and twin Q-networks. The policy was a tanh-squashed diagonal Gaussian, ensuring that sampled actions lay in the bounds of \eqref{eq:action}. Sampled actions were used during training and the squashed mean during evaluation. Standard SAC target-network, terminal-masking, critic, and entropy-regularized actor updates were used.

            The reward function used for training is shown in \eqref{eq:reward}. Here, $d_t$ is the centreline distance in millimeters from the guidewire tip to the target. In \eqref{eq:reward}, $\mathbb{I}_{\mathrm{goal},t+1}=1$ when the target was reached and zero otherwise. Thus, progress along the vessel centreline produced positive reward, movement away from the target produced negative reward, and the constant term penalized long trajectories.
            
            \begin{equation}
                r_t=-1.5\times10^{-4}+10^{-3}(d_t-d_{t+1})
                    +\mathbb{I}_{\mathrm{goal},t+1},
                \label{eq:reward}
            \end{equation}

        \subsection{Failure Predictor}

            To provide an online estimate of navigation risk, a separate recurrent predictor was trained for each navigation task using \textit{in silico} trajectories from that task. The navigation policy was trained first and subsequently used to generate successful and unsuccessful trajectories, which were stored and used to train the failure predictor \textit{a posteriori}. The predictor and navigation policy were therefore not trained jointly.
            
            At control step $t$, the predictor received the current observation $\mathbf{o}_t$, its difference from the previous observation $\Delta\mathbf{o}_t$, and normalized episode progress $\rho_t$. The predictor input and recurrent history were defined as \eqref{eq:predictor_input}, where $\Delta\mathbf{o}_0=\mathbf{0}$.

            \begin{equation}
                \begin{aligned}
                    \Delta\mathbf{o}_t
                    & = \mathbf{o}_t-\mathbf{o}_{t-1}, \\
                    \mathbf{x}_t
                    & = \left[
                    \mathbf{o}_t^\top,
                    \Delta\mathbf{o}_t^\top,
                    \rho_t
                    \right]^\top, \\
                    \mathcal{H}_t
                    & = \left(\mathbf{x}_0,\ldots,\mathbf{x}_t\right).
                \end{aligned}
                \label{eq:predictor_input}
            \end{equation}

            For each unsuccessful trajectory, the failure time step $t_f$ was defined as the first control step where the task-specific terminal failure criterion was satisfied and the episode was recorded as unsuccessful. Training labels were generated from the stored trajectory according to the temporal distance from $t_f$. The binary training label and predicted probability of failure within the subsequent $H$ time steps were defined as \eqref{eq:failure_prediction}.

            \begin{equation}
                \begin{aligned}
                    y_t &=
                    \begin{cases}
                    1, & 0<t_f-t\leq H,\\
                    0, & \text{otherwise},
                    \end{cases}\\
                    \hat q_t
                    &= \sigma!\left(f_{\phi}(\mathcal{H}_t)\right),\\
                    \hat q_t
                    &\approx
                    \Pr!\left(
                    0<t_f-t\leq H
                    \mid \mathcal{H}_t
                    \right).
                \end{aligned}
                \label{eq:failure_prediction}
            \end{equation}

            Here, $f_{\phi}$ denotes the recurrent predictor and prediction head and $\sigma(\cdot)$ is the sigmoid function. With $H=20$, the final 20 control steps preceding $t_f$ were therefore labeled $y_t=1$, while earlier time steps were labeled $y_t=0$. Successful trajectories, for which no failure time step existed, were assigned $y_t=0$ at every time step. Only trajectories generated in the training anatomy and its geometric augmentations were used for predictor development.

\begin{figure}[tb]
                \centering
                \includegraphics[width=0.7\linewidth]{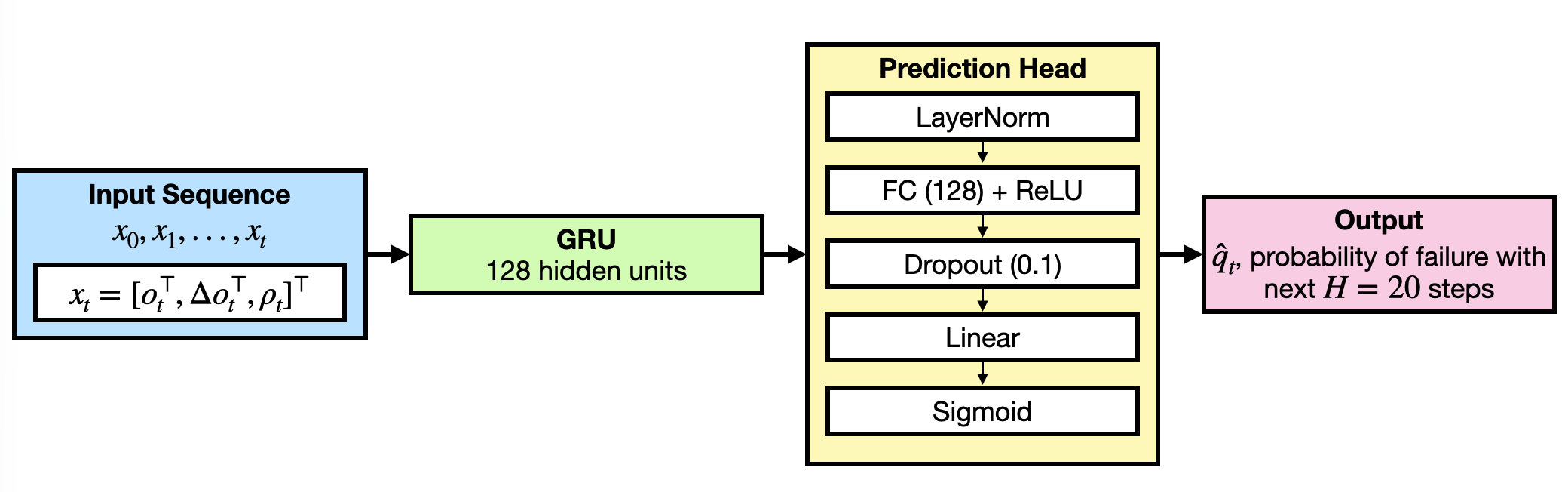}
                \caption{Failure predictor architecture, comprising the recurrent input sequence, 128-unit GRU, prediction head, and sigmoid output estimating failure probability within the next $H=20$ control steps.}
                \label{fig:failure_predictor}
            \end{figure}

            As shown in Fig.~\ref{fig:failure_predictor}, each predictor comprised a one-layer GRU with 128 hidden units followed by layer normalization, a 128-unit ReLU layer, dropout of 0.1, a linear output layer, and a sigmoid activation. The recurrent hidden state was propagated through $\mathcal{H}_t$; consequently, $H$ denotes the future prediction horizon and not the length of the GRU input history.

            The network parameters $\phi$ were learned by minimizing a class-weighted binary cross-entropy loss between the target label $y_t$ and predicted probability $\hat q_t$, as in \eqref{eq:failure_loss}.

            \begin{equation}
                \begin{aligned}
                    \mathcal{L}_{\mathrm{FP}}(\phi)
                    = -\frac{1}{N}\sum_{t=1}^{N}
                    \Big[
                    &w_1 y_t\log(\hat q_t) \\
                    &+ w_0(1-y_t)\log(1-\hat q_t)
                    \Big].
                \end{aligned}
                \label{eq:failure_loss}
            \end{equation}

            Here, $w_1$ and $w_0$ are the positive- and negative-class weights, respectively, used to compensate for class imbalance, and $N$ is the number of labeled time steps in the training batch. Models were trained using AdamW with a learning rate of $3\times10^{-4}$, weight decay of $10^{-5}$, batch size 16, and gradient clipping at 5.0. Trajectories, rather than individual time steps, were divided 70/15/15\% into training, validation, and test sets before generation of the corresponding time-step samples, preventing observations from the same episode from appearing across different data splits. The checkpoint with the lowest validation loss was selected for evaluation. 

            During deployment, the task-matched predictor operated alongside the navigation controller and updated $\hat q_t$ with each new observation. For the \textit{in vitro} experiments, an optional automatic guidewire-retraction response was implemented. When $\hat q_t>0.80$ for six consecutive control steps, the controller replaced the guidewire translation command with a normalized action of $-1.0$ for ten consecutive steps, thereby commanding guidewire withdrawal. A separate threshold of $\hat q_t\geq0.50$ generated a visual warning but did not independently initiate retraction. The mechanism was evaluated as a proof-of-concept response and was not treated as a validated safety controller.

    \subsection{Training, experimental design and evaluation}

        \subsubsection{Training}
        
            RL training was done entirely \textit{in silico}, using the hyperparameters in Table~\ref{tab:sac_hyperparameters}. All training was completed on an NVIDIA DGX A100 node (Santa Clara, CA, USA). One attempt at a navigation task constituted an episode. An episode was successful when the tracked device tip entered the predefined target region equal to the local vessel diameter. Training episodes terminated after $200$~exploration steps (approximately $27$\,s) if the target had not been reached to lower simulation cost.

            \begin{table}[tb]
                \centering
                \caption{\textit{In silico} training hyperparameters.}
                \label{tab:sac_hyperparameters}
                \begin{tabular}{lc}
                    \hline
                    Parameter & Value \\
                    \hline
                    Learning rate & $2.20\times10^{-4}$ \\
                    Discount factor ($\gamma$) & 0.99 \\
                    Batch size & 32 \\
                    Replay buffer size & $10^{4}$ \\
                    Training steps & $2\times10^{7}$ \\
                    Heat-up steps & $5\times10^{5}$ \\
                    Updates per environment step & 0.05 \\
                    Hidden layers & $4 \times 900$ \\
                    Embedder layers & 1 \\
                    Embedder width & 500 \\
                    \hline
                \end{tabular}
            \end{table}

            To select the checkpoint used for final testing, evaluations were conducted every $2.5 \times 10^5$~exploration steps for $50$~episodes during \textit{in silico} training. Agents were allowed $450$~exploration steps (equivalent to approximately $60$\,s) during \textit{in silico} evaluation to avoid classifying slower but recoverable trajectories as failures because of the training-time truncation limit. The \textit{Success Rate} was defined as the percentage of evaluation episodes in which the controller successfully reached the target; for unsuccessful episodes \textit{Path Ratio} was defined as $\mathrm{PathRatio} =100\left(1-\frac{d_{\mathrm{end}}}{d_0}\right)\%, $ where $d_0$ and $d_{\mathrm{end}}$ are the initial and final centreline distances to the target (larger values indicate greater progress); and \textit{Procedure Time} was defined as the time taken from the start of navigation to the target location for successful episodes~\cite{Robertshaw2026jaha}. Each evaluation step recorded the success rate, procedure time, and path ratio. The checkpoint with the highest success rate across training was selected for final testing. 

        \subsubsection{Experimental design \& evaluation}

\begin{figure}[tb]
                \centering
                \includegraphics[width=0.8\linewidth]{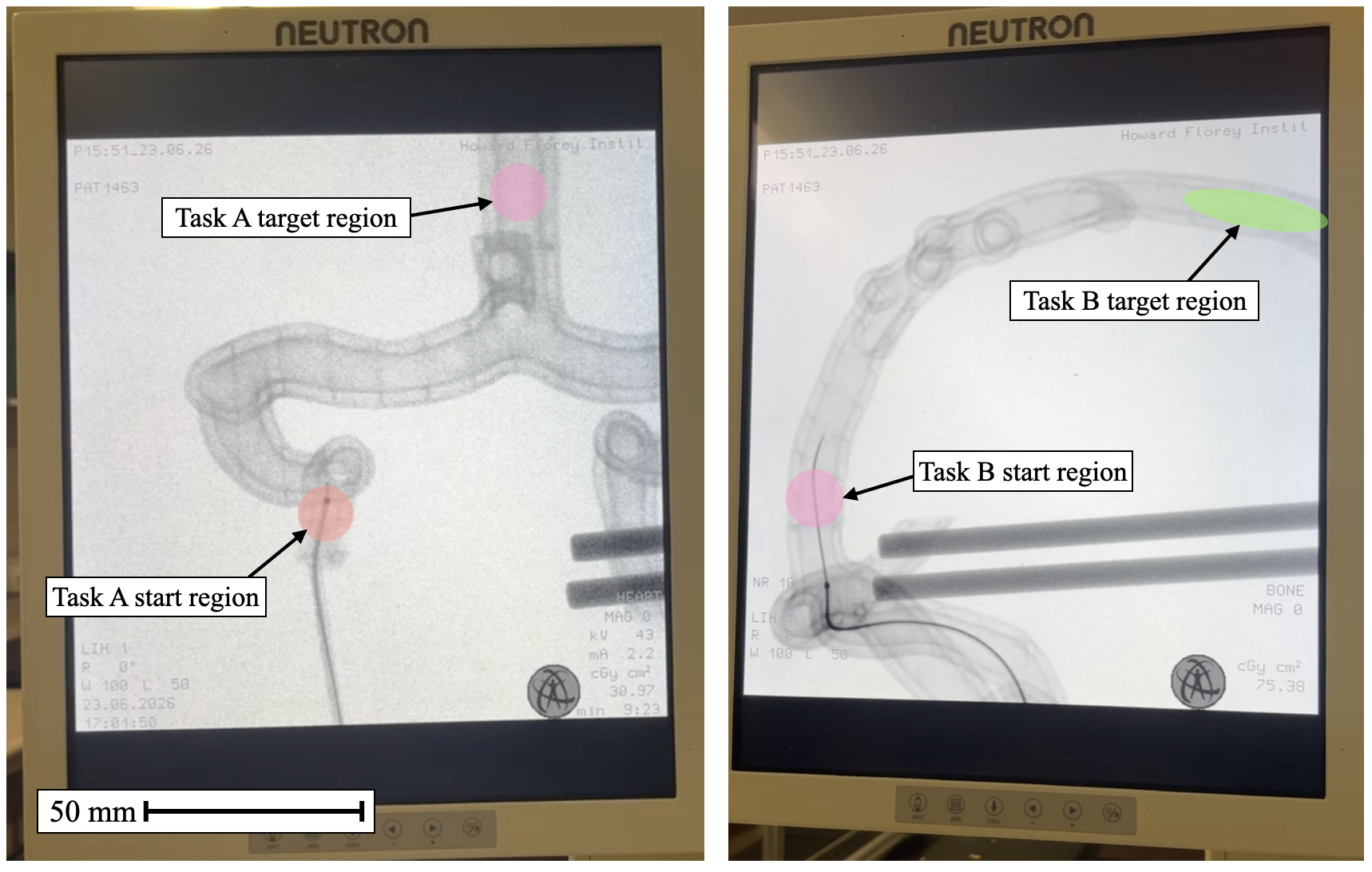}
                \caption{Phantom vessel model under fluoroscopy, showing start and target regions for task A and B.}
                \label{fig:invitrotask}
            \end{figure}

            The trained agents were evaluated separately for Tasks A and B in the training anatomy and the previously unseen hold-out anatomy (Fig.~\ref{fig:invitrotask}). Each anatomy--task condition comprised 250 \textit{in silico} episodes and five \textit{in vitro} runs. Success rate, procedure time, and path ratio were recorded for every condition. Physical runs were limited to $5$\,min to account for the lower operating speed of the robotic manipulator.

            Failure-prediction performance was evaluated at the prespecified visual-warning threshold of $\hat q_t\geq0.50$. Detection rate was the proportion of failed episodes in which $\hat q_t\geq0.50$ at least once before failure, thereby generating a warning; warning lead time was the number of control steps between the first warning and failure; and false-alarm rate was the proportion of successful episodes with at least one warning. Calibration was assessed using an episode-level Brier score, obtained by averaging $(\hat q_t-y_t)^2$ over all timesteps in an episode. This measures the accuracy of the predicted failure probabilities, with lower values indicating closer agreement between predicted risk and observed failure outcomes. Success rate, detection, and false-alarm proportions are reported with Wilson 95\% confidence intervals (CIs). Procedure time, path ratio, and Brier scores are reported as mean $\pm$ standard deviation (SD), whereas warning lead time is reported as median [interquartile range]. Given the five physical runs per anatomy--task condition, the \textit{in vitro} analysis was descriptive and no inferential between-condition tests were performed.

\section{Results}

    \subsection{\textit{In silico}}

        Agent performance across the training and hold-out anatomies is summarized in Table~\ref{tab:failure-pred-agent-performance}. In the training anatomy, Tasks A and B achieved success rates of 85.6\% and 98.4\%, respectively. In the hold-out anatomy, Task A success decreased by 43.6 percentage points to 42.0\%, whereas Task B success decreased by 6.8 percentage points to 91.6\%. Across successful episodes, mean procedure time ranged from $10.33$\,s to $13.40$\,s. Across unsuccessful episodes, mean path ratio ranged from $8.4$\% to $31.7$\%.

        \begin{table*}[t]
            \centering
            \caption{\textit{In silico} navigation performance across anatomy and task conditions. Values are reported over 250 episodes per condition. Path ratio is summarized over unsuccessful episodes, and procedure time over successful episodes. Continuous values are mean $\pm$ SD.}
            \label{tab:failure-pred-agent-performance}
            \begin{tabular}{lllcc}
            \toprule
            Anatomy & Task & Success rate (\%, 95\% CI) & Path ratio (\%) & Procedure time (s) \\
            \midrule
            Training & A & 85.6 (214/250, 80.7--89.4\%) & 13.7 $\pm$ 23.2 & 10.33 $\pm$ 4.60 \\
            Training & B & 98.4 (246/250, 96.0--99.4\%) & 31.7 $\pm$ 4.5 & 13.40 $\pm$ 5.64 \\
            Hold-out & A & 42.0 (105/250, 36.0--48.2\%) & 8.4 $\pm$ 1.4 & 12.15 $\pm$ 6.44 \\
            Hold-out & B & 91.6 (229/250, 87.5--94.4\%) & 30.2 $\pm$ 19.3 & 11.77 $\pm$ 5.74 \\
            \bottomrule
            \end{tabular}
        \end{table*}

        Quantitative failure-prediction results are reported in Table~\ref{tab:failure-pred-matched-general}, where the predictor detected every failure in the training anatomy. In the hold-out anatomy, the predictor was able to detect $99.3$\% of Task A failures and $100.0$\% of Task B failures. Median warning lead time ranged from $22$ to $27$ steps, corresponding to approximately $2.9$--$3.6$\,s at the \textit{in silico} evaluation control rate. Maximum false-alarm rates were at most $2.3$\% in the training anatomy and at most $6.7$\% in the hold-out anatomy. The largest calibration difference occurred for hold-out Task A, for which the Brier score was $0.0203$.

        \begin{table*}[t]
            \centering
            \setlength{\tabcolsep}{3pt}
            \caption{Failure-predictor performance \textit{in silico}. Proportion confidence intervals are Wilson 95\% CIs. Warning lead time is median [IQR], and Brier score is mean $\pm$ SD.}
            \label{tab:failure-pred-matched-general}
            \begin{tabular}{llcccc}
                \toprule
                Anatomy & Task & Detection (\%, 95\% CI) & Lead time (steps) & False alarm (\%, 95\% CI) & Brier score \\
                \midrule
                Training & A & 100.0 (36/36, 90.4--100.0\%) & 23.0 [21.8, 24.0] & 2.3 (5/214, 1.0--5.4\%) & $0.0032 \pm 0.0122$ \\
                Training & B & 100.0 (4/4, 51.0--100.0\%) & 22.0 [21.5, 22.5] & 0.8 (2/246, 0.2--2.9\%) & $0.0003 \pm 0.0018$ \\
                Hold-out & A & 99.3 (144/145, 96.2--99.9\%) & 27.0 [26.0, 28.0] & 6.7 (7/105, 3.3--13.1\%) & $0.0203 \pm 0.0212$ \\
                Hold-out & B & 100.0 (21/21, 84.5--100.0\%) & 22.0 [20.0, 29.0] & 0.9 (2/229, 0.2--3.1\%) & $0.0026 \pm 0.0103$ \\
                \bottomrule
            \end{tabular}
        \end{table*}

\begin{figure}[t]
            \centering
            \includegraphics[width=0.9\linewidth]{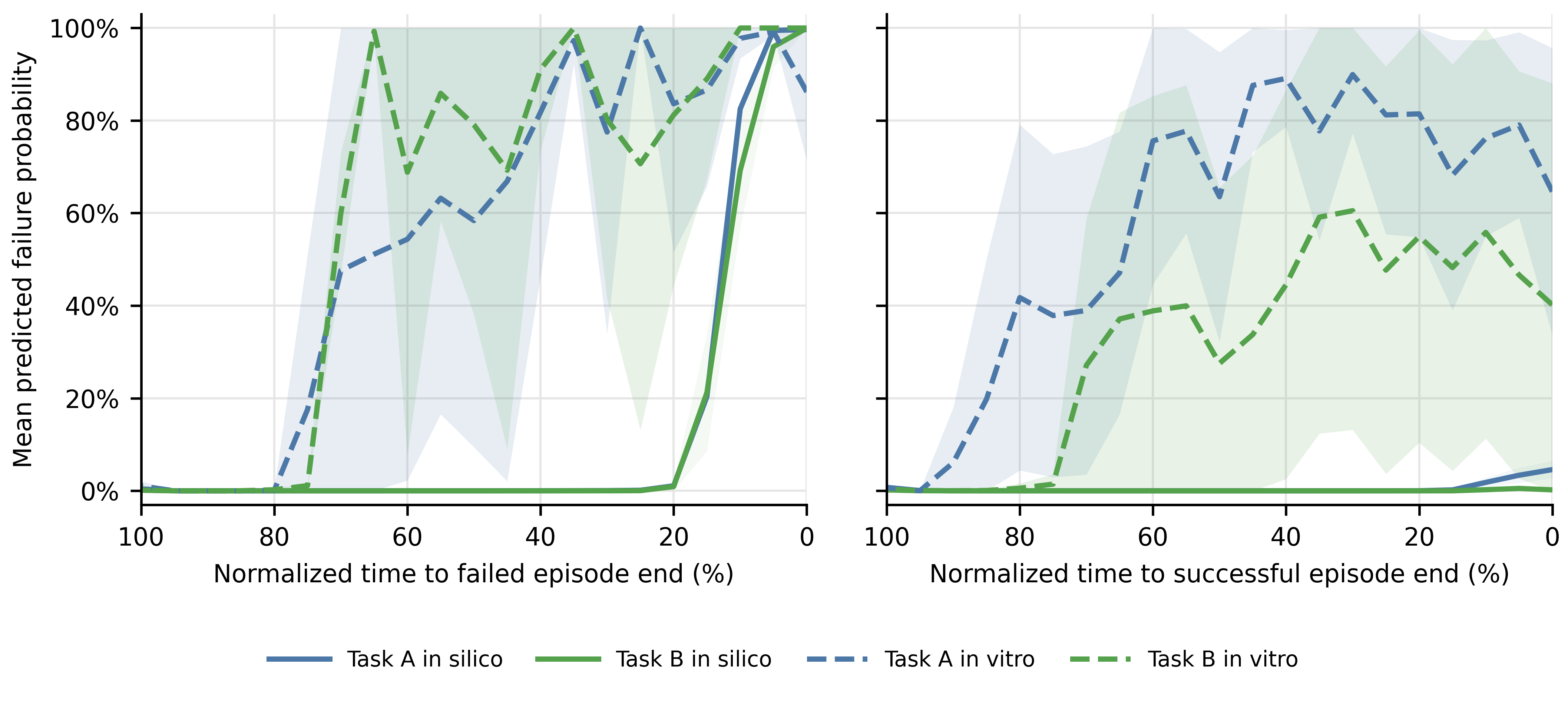}
            \caption{Temporal failure-prediction performance, across training and hold-out anatomies. Mean predictor outputs are shown for unsuccessful episodes aligned to failure (left) and successful episodes aligned to completion (right). Time is normalized to episode duration. Solid and dashed curves are \textit{in silico} and \textit{in vitro} episodes, respectively; shaded regions show 95\% CIs.}
            \label{fig:failure-time}
        \end{figure}

        The solid curves in Figure~\ref{fig:failure-time} show the temporal outputs of the task-matched predictors in simulation. For unsuccessful episodes, the mean predicted probability remained low during the early and middle portions of the trajectory and rose sharply as failure approached, reaching approximately $100$\% near episode termination for both tasks. For successful episodes, the mean output remained close to zero throughout, consistent with the low \textit{in silico} false-alarm rates in Table~\ref{tab:failure-pred-matched-general}.

    \subsection{\textit{In vitro}}

        The \textit{in vitro} navigation results are summarized in Table~\ref{tab:in-vitro-navigation-summary}. Success occurs in 14/20 runs, corresponding to an overall success rate of 70\% (95\% CI 48.1--85.5\%). In the training anatomy, both Tasks A and B achieved success rates of 80\%, while in the hold-out anatomy, Task A achieved a 40\% success rate and Task B achieved 80\%.
        
        For successful runs in the training anatomy, mean procedure time ranged from $85$\,s for Task B to $189$\,s for Task A. Among unsuccessful runs, Task A had a higher path ratio than Task B in both anatomies. In the training anatomy, the single unsuccessful Task A run reached $84$\% of the route, whereas unsuccessful Task B runs reached $35$\%. In the hold-out anatomy, the corresponding values were $88$\% for Task A and $56$\% for Task B.
        
        The observed physical failure modes differed by task and anatomy. For Task~A, the single training-anatomy failure occurred when the device entered the left internal jugular vein instead of the SSS. In the hold-out anatomy, one Task~A failure occurred when the device became lodged in the right internal jugular vein and two occurred at the base of the SSS. For Task~B, the single failure in each anatomy occurred when the device became lodged in a smaller branch arising from the SSS. All hold-out anatomy failures were associated with substantial device–phantom friction.
        
        \begin{table}[t]
            \centering
            \caption{\textit{In vitro} navigation performance across anatomy and task conditions. Each condition comprised five runs. Path ratio is summarized over unsuccessful runs, and procedure time over successful runs. Continuous values are mean $\pm$ SD.}
            \label{tab:in-vitro-navigation-summary}
            \renewcommand{\arraystretch}{1.1}
            \begin{tabular}{llccc}
                \toprule
                Anatomy & Task & Success rate (\%) & Path ratio (\%) & Procedure time (s) \\
                \midrule
                Training & A & 80 (4/5) & $84$ ($n=1$) & $188 \pm 96$ \\
                Training & B & 80 (4/5) & $35$ ($n=1$) & $85 \pm 29$ \\
                Hold-out & A & 40 (2/5) & $88 \pm 10$ & $101 \pm 33$ \\
                Hold-out & B & 80 (4/5) & $56$ ($n=1$) & $93 \pm 56$ \\
                \bottomrule
            \end{tabular}
        \end{table}

\begin{figure}[t]
            \centering
            \includegraphics[width=0.8\linewidth]{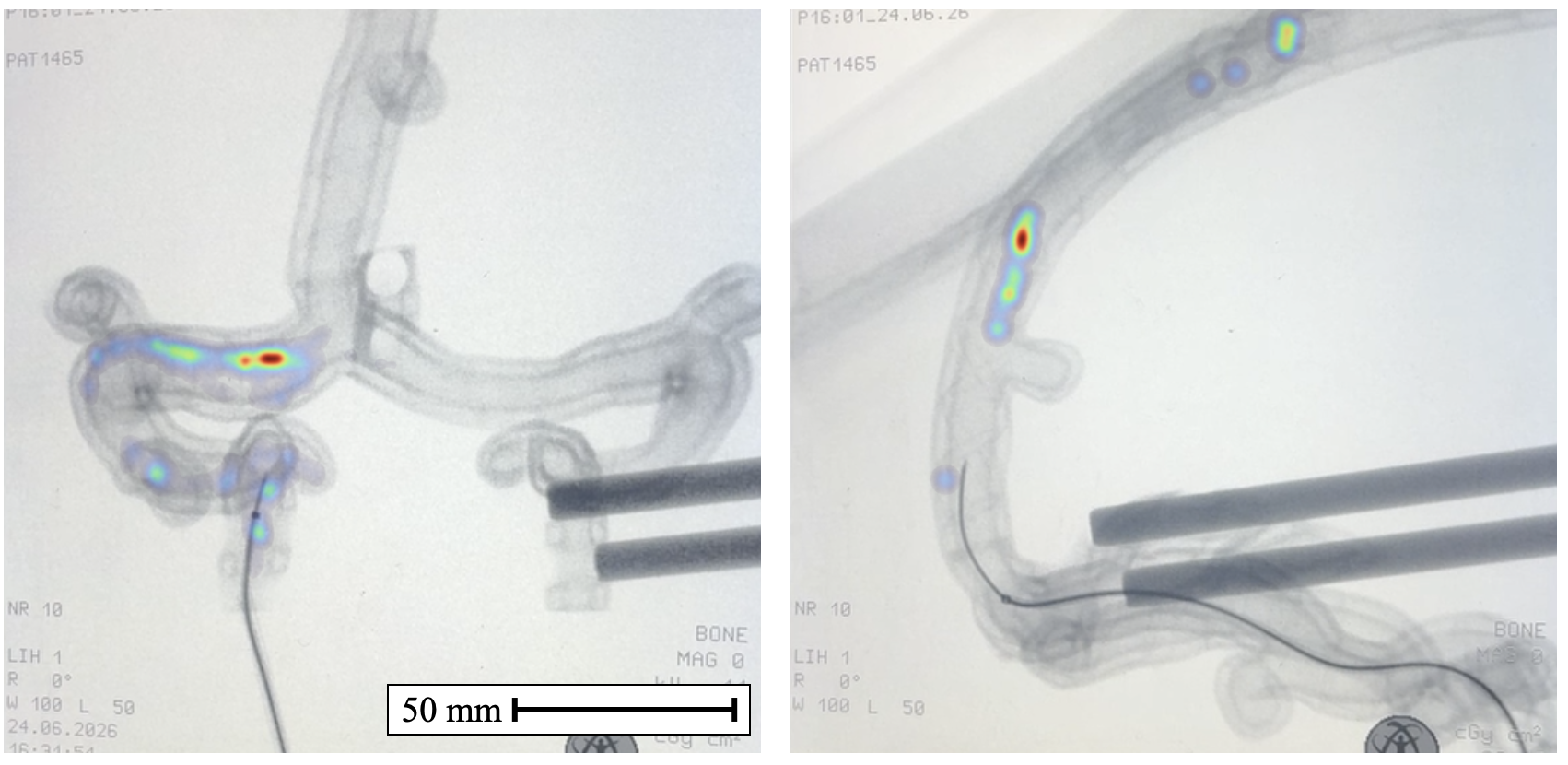}
            \caption{Spatial distribution of predictor-triggered guidewire-retraction onsets during armed navigation in the hold-out \textit{in vitro} anatomy. Heatmap intensity indicates the relative frequency of retraction onset.}
            \label{fig:retraction-heatmap}
        \end{figure}

        The dashed curves in Figure~\ref{fig:failure-time} show task-matched predictor outputs during \textit{in vitro} testing. Mean predicted risk increased before failed episodes for both tasks and was high near failure. In contrast to \textit{in silico}, the physical predictor outputs were also elevated during successful runs, as Task A mean exceeded $50$\% through much of the latter portion of successful trajectories, while Task B mean rose to approximately 40--60\%. The wide CIs reflect the small number of physical runs, but the overlap between successful and unsuccessful outputs indicates reduced predictor calibration after sim-to-real transfer. Figure~\ref{fig:retraction-heatmap} maps the locations at which the armed automatic response initiated guidewire retraction in the hold-out \textit{in vitro} phantom. 

\section{Discussion}

    This study presents the first demonstration of autonomous robotic navigation in the cerebral venous system and evaluates a complementary online failure predictor. Two sequential navigation tasks approximating the access route from the superior right jugular vein to the SSS were completed \textit{in silico} and \textit{in vitro} under fluoroscopic guidance, including a hold-out anatomy. The primary findings were that autonomous venous access was feasible but strongly task- and anatomy-dependent; that simulated failure prediction generalized well to the hold-out anatomy; and that direct transfer of the predictor to the physical testbed introduced calibration error. The work therefore establishes technical feasibility while identifying the limitations that must be addressed before autonomous BCI delivery can be considered safe or clinically applicable.

    \subsection{Autonomous navigation performance}

        The navigation results demonstrate that cross-anatomy generalization cannot be characterized by a single aggregate success rate. Task B retained high \textit{in silico} performance in the hold-out anatomy, decreasing from $98.4$\% to $91.6$\%, whereas Task A decreased from $85.6$\% to $42.0$\%. Task A requires route selection at multiple venous junctions, with greater path tortuosity than Task B, making it a more challenging navigation. The larger degradation for Task A suggests that the geometric augmentation used during training did not span the variation in this junction sufficiently. It also shows why validation confined to a training anatomy would have overstated the robustness of the controller.

        \textit{In vitro} navigation was successful in 14/20 runs. The hold-out results followed the same task-dependent pattern as simulation: Task A achieved $40$\% physical success compared with $42$\% in simulation, whereas Task B achieved $80$\% compared with $91.6$\%. These similar point estimates should not be interpreted as evidence of sim-to-real equivalence because the physical estimates were based on only five runs per condition and consequently had wide CIs. Nevertheless, successful \textit{in vitro} fluoroscopy-guided navigation in both printed anatomies shows that a policy trained entirely \textit{in silico} can produce useful physical behavior in the cerebral venous route, extending prior autonomous endovascular work from arterial navigation toward venous BCI access~\cite{Robertshaw2025ipcai,Robertshaw2026ral,Robertshaw2026iros}.

        The \textit{in vitro} failure modes identify specific targets for system improvement. Incorrect entry into the contralateral jugular route during Task A may reflect state ambiguity or an incorrect policy decision at a vascular branching point, whereas lodging at the SSS base or in smaller SSS branches was attributed to friction. The latter failures expose a limitation of matching \textit{in silico} device mechanics to a rigid printed phantom, as contact and friction can dominate progression even when the selected route is anatomically correct. \textit{In vitro} procedure times were also longer than \textit{in silico} procedure times, due to the slower robotic platform, and fluoroscopic acquisition. These differences reinforce the need to evaluate task completion, mechanics, and timing after \textit{in vitro} transfer rather than relying on \textit{in silico} performance alone.

        For endovascular BCIs, reaching the SSS is an enabling step rather than the complete implantation problem. The optimal recording location depends on the relationship among venous anatomy, task-relevant cortex, and electrode coverage~\cite{Schone2025}. Future clinical systems would need to register a patient-specific implantation target derived from venography and functional imaging, navigate to that target, and support controlled device deployment. The present experiments address autonomous access only, as no Stentrode was deployed, no neural signals were recorded, and the accuracy of final electrode placement relative to functional cortex was not evaluated.

    \subsection{Failure prediction and human oversight}

        \textit{In silico} task-specific predictors detected 99.3--100.0\% of failures with median warning lead times of approximately 20--27 control steps. The \textit{in vitro} trajectories provide a more cautious result, as predicted risk rose before failed runs, showing that the learned temporal signal retained some relevance after transfer. However, risk also rose during successful \textit{in vitro} runs and frequently exceeded the 50\% visual-warning threshold, especially for Task A. This behavior is consistent with a covariate shift in tracked device motion, timing, or image-derived coordinates between \textit{in silico} and \textit{in vitro} testbeds. A detector that identifies failures but also warns during successful navigation could increase operator workload or cause unnecessary withdrawal, limiting its clinical value. Physical calibration, task-conditional threshold selection, and explicit uncertainty or out-of-distribution detection are therefore necessary before the output can be used as a dependable supervisory signal.

        Coupling persistent high predicted risk to guidewire retraction illustrates how online monitoring could lead to an immediate physical response. However, it does not establish that automatic retraction is safe or beneficial. No matched no-retraction experiments were performed, and retraction could itself prolong a procedure, cause repeated device--wall contact, or prevent recovery from a transient but non-hazardous state. Near-term use should therefore remain operator-armed and human-supervised, consistent with proposals for AI assistance within existing interventional workflows~\cite{Robertshaw2026jaha,Granados2026}.

    \subsection{Limitations and future work}

        Several limitations should be acknowledged. First, only two venous anatomies were available, with one used for training and one for hold-out evaluation, reflecting the limited availability of cerebral venous imaging datasets in the current literature, when compared to the corresponding availability of arterial vascular datasets. Geometric augmentation increases variation around the training model but cannot reproduce the topology and pathology of a larger patient cohort. This limited-anatomy design is consistent with earlier feasibility studies in autonomous endovascular navigation, in which initial single-anatomy demonstrations motivated subsequent evaluation across larger patient-derived cohorts~\cite{Robertshaw2024}. The \textit{in vitro} study included only five runs per anatomy--task condition, limiting precision and preventing meaningful inferential comparisons. No manual expert-navigation benchmark was included, preventing direct comparison of autonomous and operator performance in terms of success rate, procedure time, or failure modes. Future studies should use a larger set of patient-derived venographies, preserve a multi-patient external test cohort, and repeat \textit{in vitro} evaluation sufficiently to characterize both common and rare failures.

        Second, 3D-printed phantoms did not reproduce pulsatility, or physiological motion~\cite{Robertshaw2026jaha}, although these are less prevalent in the venous system when compared to the arterial system. The frictional behavior was a major cause of failure and may not represent either living veins or more compliant flow phantoms. More realistic flow loops, compliant materials, and systematic domain randomization of image and device mechanics are required to enhance the \textit{in vitro} testbed. Future testbeds should therefore investigate elastomeric materials that better reproduce venous deformation and device--wall interaction, with physiologically representative flow conditions.

        Third, navigation success rate was defined by target entry and did not measure contact force, vessel-wall stress, perforation, dissection, or other safety outcomes. It also quantified vascular access rather than optimal BCI placement relative to the intended cortical target and electrode coverage. No endovascular BCI was deployed, and no \textit{ex vivo} or \textit{in vivo} validation was performed. Future platforms should incorporate force estimation, impose explicit mechanical safety constraints, evaluate final implant-placement accuracy relative to patient-specific functional targets, evaluate controlled catheter and implant deployment, and progress through compliant phantoms and preclinical models before any clinical study.

        Finally, the predictors were trained from trajectories generated by the same controller family and may not recognize failures caused by unseen policies, tracking artifacts, or hardware faults. \textit{In vitro} results show that simulation calibration did not transfer directly. Future work should fine-tune and recalibrate the predictor using independently collected physical data, evaluate generalization across controllers and failure modes, prospectively select alert thresholds, and compare operator-only, warning-only, and automatic-response conditions. A controlled study is also needed to determine whether prediction and retraction improve success, reduce force, or shorten recovery without increasing false interventions.
    
\section{Conclusion}

    This work demonstrates the feasibility of autonomous robotic access to the cerebral venous system for an endovascular BCI route. \textit{In silico}-trained controllers completed two sequential tasks from the right jugular vein toward the SSS in training and hold-out anatomies and achieved 70\% success across 20 fluoroscopy-guided physical runs. Recurrent predictors identified nearly all simulated failures with early warning and low false-alarm rates. However, elevated risk estimates during successful \textit{in vitro} runs revealed an important sim-to-real calibration gap. The results support further development of human-supervised autonomous venous navigation, but they do not establish autonomous implantation or clinical safety. Larger anatomical cohorts, realistic compliant flow models, physical recalibration, force-aware safety evaluation, and preclinical validation are required before translation.

\section*{Acknowledgments}

Partial financial support was received from the WELLCOME TRUST (Grant Agreement No 203148/A/16/Z), the Engineering and Physical Sciences Research Council Doctoral Training Partnership (Grant Agreement No EP/R513064/1), and the MRC IAA 2021 King's College London (MR/X502923/1). For the purpose of Open Access, the Author has applied a CC BY public copyright license to any Author Accepted Manuscript version arising from this submission.

\bibliographystyle{unsrt}  
\bibliography{references}

@article{Chaudhary2016,
  author = {Chaudhary, Ujwal and Birbaumer, Niels and Ramos-Murguialday, Aitor},
  title = {Brain--computer interfaces for communication and rehabilitation},
  journal = {Nature Reviews Neurology},
  year = {2016},
  volume = {12},
  number = {9},
  pages = {513--525},
  doi = {10.1038/nrneurol.2016.113}
}

@Article{NicolasAlonso2012,
AUTHOR = {Nicolas-Alonso, Luis Fernando and Gomez-Gil, Jaime},
TITLE = {Brain Computer Interfaces, a Review},
JOURNAL = {Sensors},
VOLUME = {12},
YEAR = {2012},
NUMBER = {2},
PAGES = {1211--1279},
PubMedID = {22438708},
ISSN = {1424-8220},
DOI = {10.3390/s120201211}
}

@article{Jackson2012,
  author = {Jackson, Andrew and Zimmermann, Jakob B.},
  title = {Neural interfaces for the brain and spinal cord---restoring motor function},
  journal = {Nature Reviews Neurology},
  year = {2012},
  volume = {8},
  number = {12},
  pages = {690--699},
  doi = {10.1038/nrneurol.2012.219}
}

@article{PatrickKrueger2025,
  author = {Patrick-Krueger, K. Michelle and Burkhart, Ian and Contreras-Vidal, Jose L.},
  title = {The state of clinical trials of implantable brain--computer interfaces},
  journal = {Nature Reviews Bioengineering},
  year = {2025},
  volume = {3},
  pages = {50--67},
  doi = {10.1038/s44222-024-00239-5}
}

@article{Salatino2017,
  author = {Salatino, Joseph W. and Ludwig, Kip A. and Kozai, Takashi D. Y. and Purcell, Erin K.},
  title = {Glial responses to implanted electrodes in the brain},
  journal = {Nature Biomedical Engineering},
  year = {2017},
  volume = {1},
  pages = {862--877},
  doi = {10.1038/s41551-017-0154-1}
}

@article{Campbell2018,
  author = {Campbell, Andrew and Wu, Chengyuan},
  title = {Chronically Implanted Intracranial Electrodes: Tissue Reaction and Electrical Changes},
  journal = {Micromachines},
  year = {2018},
  volume = {9},
  number = {9},
  pages = {430},
  doi = {10.3390/mi9090430}
}

@article{Ognard2025,
  author  = {Ognard, Julien and others},
  title   = {Advances in endovascular brain computer interface: Systematic review and future implications},
  journal = {Journal of Neuroscience Methods},
  volume  = {420},
  pages   = {110471},
  year    = {2025},
  doi     = {10.1016/j.jneumeth.2025.110471}
}

@article{Schone2025,
	author = {Schone, Hunter R. and others},
	title = {Motor Cortex Coverage Predicts Signal Strength of a Stentrode Endovascular Brain-Computer Interface},
	elocation-id = {2025.09.19.25335875},
	year = {2025},
	doi = {10.1101/2025.09.19.25335875},
	publisher = {Cold Spring Harbor Laboratory Press},
	journal = {medRxiv}
}

@article{Chetty2025,
	author = {Chetty, Nikole and others},
	title = {Signal properties and stability of a chronically implanted endovascular brain computer interface},
	elocation-id = {2025.09.19.25335897},
	year = {2025},
	doi = {10.1101/2025.09.19.25335897},
	publisher = {Cold Spring Harbor Laboratory Press},
	journal = {medRxiv}
}

@article{Qi2024,
  author = {Qi, Weijie and Ooi, Andrew and Grayden, David B. and Opie, Nicholas L. and John, Sam E.},
  title = {Haemodynamics of stent-mounted neural interfaces in tapered and deformed blood vessels},
  journal = {Scientific Reports},
  year = {2024},
  volume = {14},
  pages = {7212},
  doi = {10.1038/s41598-024-57460-w}
}

@article{Robertshaw2023,
  author  = {Robertshaw, H. and others},
  title   = {Artificial intelligence in the autonomous navigation of endovascular interventions: a systematic review},
  journal = {Frontiers in Human Neuroscience},
  volume  = {17},
  pages   = {1239374},
  year    = {2023},
  doi     = {10.3389/fnhum.2023.1239374},
}

@ARTICLE{Robertshaw2026ral,
  author={Robertshaw, H. and others},
  journal={IEEE Robotics and Automation Letters}, 
  title={Toward AI Autonomous Navigation for Mechanical Thrombectomy Using Hierarchical Modular Multi-Agent Reinforcement Learning (HM-MARL)}, 
  year={2026},
  volume={11},
  number={4},
  pages={4745-4752},
  doi={10.1109/LRA.2026.3664661}}

@article{Robertshaw2025ipcai,
  author  = {Robertshaw, H. and others},
  title   = {Reinforcement learning for safe autonomous two-device navigation of cerebral vessels in mechanical thrombectomy},
  journal = {Int J CARS},
  volume  = {20},
  pages   = {1077--1086},
  year    = {2025},
  doi     = {10.1007/s11548-025-03339-8},
}

@misc{Robertshaw2026iros,
      title={Toward Safe Autonomous Robotic Endovascular Interventions using World Models}, 
      author={Harry Robertshaw and others},
      year={2026},
      eprint={2604.20151},
      archivePrefix={arXiv},
      primaryClass={cs.RO},
      url={https://arxiv.org/abs/2604.20151}, 
}

@InProceedings{Robertshaw2025miccai,
    author = {Robertshaw, Harry and others},
    title = {World Model for {AI} Autonomous Navigation in Mechanical Thrombectomy},
    booktitle = {Medical Image Computing and Computer Assisted Intervention 2025},
    year = {2025},
    page = {683 -- 693}
}

@article{Robertshaw2026jaha,
author = {Harry Robertshaw and others},
title = {A Position Statement on Endovascular Models and Effectiveness Metrics for Mechanical Thrombectomy Navigation, on Behalf of the Stakeholder Taskforce for Artificial Intelligence\&\#x2013;Assisted Robotic Thrombectomy (START)},
journal = {Journal of the American Heart Association},
volume = {15},
number = {7},
pages = {e044931},
year = {2026},
doi = {10.1161/JAHA.125.044931},
}

@article{Granados2026,
  author  = {Granados, A. and others},
  title   = {Evolving surgical teams in the age of artificial intelligence and robotics},
  journal = {Frontiers in Science},
  volume  = {4},
  pages   = {1783803},
  year    = {2026},
  doi     = {10.3389/fsci.2026.1783803},
}

@article{Fedorov2012,
   author = {Andriy Fedorov and others},
   doi = {10.1016/j.mri.2012.05.001},
   issn = {0730725X},
   issue = {9},
   journal = {Magnetic Resonance Imaging},
   month = {11},
   pages = {1323-1341},
   pmid = {22770690},
   title = {3D Slicer as an image computing platform for the Quantitative Imaging Network},
   volume = {30},
   year = {2012},
}

@book{Faure2012,
   author = {François Faure and others},
   city = {Berlin, Heidelberg},
   isbn = {978-3-642-29014-5},
   pages = {283-321},
   publisher = {Springer Berlin Heidelberg},
   title = {SOFA, a Multi-Model Framework for Interactive Physical Simulation},
   year = {2012},
}

@inbook{Wei2012,
   author = {Yiyi Wei and Stephane Cotin and Jrmie Dequidt and Christian Duriez and Jrmie Allard and Erwan Kerrie},
   doi = {10.5772/48635},
   journal = {Aneurysm},
   month = {8},
   publisher = {InTech},
   title = {A (Near) Real-Time Simulation Method of Aneurysm Coil Embolization},
   year = {2012},
}

@article{Karstensen2025,
   author = {Lennart Karstensen and others},
   doi = {10.1016/j.compbiomed.2025.110844},
   issn = {00104825},
   journal = {Computers in Biology and Medicine},
   month = {9},
   pages = {110844},
   title = {Learning-based autonomous navigation, benchmark environments and simulation framework for endovascular interventions},
   volume = {196},
   year = {2025}
}

@article{Karstensen2022,
   author = {Lennart Karstensen and others},
   doi = {10.1007/s11548-022-02646-8},
   issn = {18616429},
   issue = {11},
   journal = {Int J CARS},
   month = {11},
   pages = {2033-2040},
   pmid = {35604490},
   publisher = {Springer Science and Business Media Deutschland GmbH},
   title = {Learning-based autonomous vascular guidewire navigation without human demonstration in the venous system of a porcine liver},
   volume = {17},
   year = {2022},
}

@article{Robertshaw2024,
   author = {Harry Robertshaw and Lennart Karstensen and Benjamin Jackson and Alejandro Granados and Thomas C. Booth},
   doi = {10.1007/s11548-024-03208-w},
   issn = {1861-6429},
   journal = {Int J CARS},
   month = {6},
   title = {Autonomous navigation of catheters and guidewires in mechanical thrombectomy using inverse reinforcement learning},
   year = {2024}
}

@article{Oxley2016,
  author  = {Oxley, T. J. and others},
  title   = {Minimally invasive endovascular stent-electrode array for high-fidelity, chronic recordings of cortical neural activity},
  journal = {Nature Biotechnology},
  volume  = {34},
  pages   = {320--327},
  year    = {2016},
  doi     = {10.1038/nbt.3428},
}

@inproceedings{Haarnoja2018,
   author = {Tuomas Haarnoja and others},
   title = {Soft Actor-Critic: Off-Policy Maximum Entropy Deep Reinforcement Learning with a Stochastic Actor},
   year = {2018},
   booktitle = {Proceedings of the 35th International Conference on Machine Learning},
}

@inproceedings{Eyberg2022,
   author = {Christoph Eyberg and others},
   doi = {10.1515/cdbme-2022-0023},
   issn = {23645504},
   issue = {1},
   booktitle = {Current Directions in Biomedical Engineering},
   month = {7},
   pages = {89-92},
   publisher = {Walter de Gruyter GmbH},
   title = {A {ROS2}-based Testbed Environment for Endovascular Robotic Systems},
   volume = {8},
   year = {2022},
}

@article{Barnes2025,
  title={Toward Autonomous Soft Robotic Endovascular Navigation via Imitation Learning},
  author={Barnes, Noah and others},
  journal={arXiv preprint arXiv:2510.09497},
  year={2025},
  doi={10.48550/arXiv.2510.09497},
  url={https://arxiv.org/abs/2510.09497}
}

@misc{DilmenCNSVenography,
  author       = {Nevit Dilmen},
  title        = {{CNS Venography 3D SR Nevit Dilmen.stl}},
  year         = {2017},
  howpublished = {Wikimedia Commons},
  url          = {https://commons.wikimedia.org/wiki/File:CNS_Venography_3D_SR_Nevit_Dilmen.stl}
}

@misc{Dilmen3DPX,
  author       = {Nevit Dilmen},
  title        = {{3DPX-003440 CNS Venography Nevit Dilmen.stl}},
  year         = {2017},
  howpublished = {Wikimedia Commons},
  url          = {https://commons.wikimedia.org/wiki/File:3DPX-003440_CNS_Venography_Nevit_Dilmen.stl}
}

\end{document}